%% file: main.tex
\pdfoutput=1

\documentclass[11pt]{article}

\usepackage[final]{acl}

\usepackage{times}
\usepackage{latexsym}

\usepackage[T1]{fontenc}

\usepackage[utf8]{inputenc}

\usepackage{microtype}

\usepackage{inconsolata}

\usepackage{graphicx}

\usepackage[utf8]{inputenc} 
\usepackage[T1]{fontenc}    
\usepackage{hyperref}       
\usepackage{url}            
\usepackage{booktabs}       
\usepackage{amsfonts}       
\usepackage{nicefrac}       
\usepackage{microtype}      
\usepackage{lipsum}		
\usepackage{graphicx}
\usepackage{natbib}
\usepackage{doi}
\usepackage{adjustbox}   
\usepackage{booktabs}    
\usepackage{array}       

\usepackage{cuted}
\usepackage{enumitem}
\usepackage{xcolor}      
\usepackage{tcolorbox}   
\usepackage{colortbl}
\usepackage{geometry}
\usepackage{CJKutf8}
\usepackage{multirow}
\usepackage{multicol}
\usepackage{dblfloatfix}
\usepackage{listings}
\usepackage{listingsutf8}

\title{More Data or Better Data? A Critical Analysis of Data Selection and Synthesis for Mathematical Reasoning}

\author{
Yike Zhao\textsuperscript{1}, 
Simin Guo\textsuperscript{2}, 
Ziqing Yang\textsuperscript{3}, 
Shifan Han\textsuperscript{3}, 
Dahua Lin\textsuperscript{4}, 
Fei Tan\textsuperscript{1}\thanks{Corresponding Author.} \\
\textsuperscript{1}East China Normal University
\\
\textsuperscript{2}University of Chicago\\
\textsuperscript{3}Independent Researcher\\
\textsuperscript{4}The Chinese University of Hong Kong\\
\texttt{ykzhao@stu.ecnu.edu.cn, simin\_guo2885@163.com, ftan@mail.ecnu.edu.cn} \\
}

\begin{document}
\maketitle
\begin{abstract}
The reasoning capabilities of Large Language Models (LLMs) play a critical role in many downstream tasks, yet depend strongly on the quality of training data. Despite various proposed data construction methods, their practical utility in real-world pipelines remains underexplored.  In this work, we conduct a comprehensive analysis of open-source datasets and data synthesis techniques for mathematical reasoning, evaluating them under a unified pipeline designed to mirror training and deployment scenarios. We further distill effective data selection strategies and identify practical methods suitable for industrial applications. Our findings highlight that structuring data in more interpretable formats, or distilling from stronger models often outweighs simply scaling up data volume. This study provides actionable guidance for integrating training data to enhance LLM capabilities, supporting both cost-effective data curation and scalable model enhancement. We hope this work will inspire further research on how to balance "more data" versus "better data" for real-world reasoning tasks.
\end{abstract}

\input{intro}
\input{selection}
\input{pretrain}

\input{finetune}
\input{attempts}
\input{future_work}

\input{conclusion}

\section*{Limitations}

Although our work summarizes practical experiences in data selection and synthesis, there are still several limitations that should be acknowledged. Our evaluation and analysis mainly focus on the mathematical reasoning capabilities of models, and the insights are drawn from representative methods and datasets. However, the scope of our study is limited, which may not fully generalize to other settings and future research. Another limitation is that the base models used in our experiments have relatively small parameter sizes. This choice helps ensure feasible training and reproducibility in industrial settings, but may underestimate the effectiveness of data synthesis methods for larger models. Besides, in industrial workflows, statistical validation (e.g., confidence intervals, p-values) is often challenging to implement due to scale and computational constraints. Therefore, we  complement performance metrics with case studies to validate the approach's effectiveness instead. We leave more comprehensive study across boarder tasks and larger model scales for future work.

\bibliography{custom}

\input{appendix}

\end{document}

%% file: intro.tex
\section{Introduction}
\label{sec:intro}

High-quality training data is widely recognized as a key factor in improving model performance across various machine learning and NLP tasks. 
With the rapid development of advanced large language models (LLMs), a growing number of high-quality synthetic datasets and domain-specific data generation methods have been developed based on LLMs \citep{tan2024largelanguagemodelsdata, wang2024surveydatasynthesisaugmentation, zhou2024surveydataaugmentationlarge, ding2024dataaugmentationusinglarge, xu2024magpiealignmentdatasynthesis,
ziegler2024craftdatasettaskspecificsynthetic, riaz2025metasynthmetapromptingdrivenagenticscaffolds, vanherle2025cutandsplatleveraginggaussiansplatting,
mathsmith,
wang2024rewarddifferenceoptimizationsample}.

However, most of these methods focus on theoretical performance in academic research, rather than practical model development in industrial contexts. In industrial practice, models often need to handle diverse tasks simultaneously, and integrating heterogeneous datasets can lead to conflicting objectives and degraded performance \citep{li2021dynamictransfermultisourcedomain, Yang_2023, yu2020gradientsurgerymultitasklearning, pike, zhang-etal-2024-balancing}. This raises a new question: \textit{Is simply adding more data always beneficial, or should we prioritize better and more targeted data?} 

In this work, we investigate the effectiveness of data selection strategies and synthesis methods in the math domain, with a focus on model development in industrial contexts.
First, we \textbf{evaluate existing open-source datasets} and \textbf{distill data selection strategies} based on the analysis. Then, we \textbf{systematically analyze data synthesis methods within a unified framework}, covering both \textit{pretraining data refinement} and \textit{supervised fine-tuning} (SFT) \textit{data generation}.
We report the implementation results of some methods, along with insights and observations from an industrial standpoint. 
Furthermore, we \textbf{report several unsuccessful attempts}, which may offer insights that are valuable as those from successful approaches.
Finally, we \textbf{propose some promising directions for future work}, including RL-inspired data synthesis techniques, to enhance controllability and effectiveness in real-world settings.

Our main contributions are as follows:

\begin{itemize}
[topsep=0pt,parsep=1pt,leftmargin=1.5em]
\item We adopt a unified evaluation pipeline that is designed to closely mirror both training and deployment scenarios, enabling realistic assessment of data effectiveness in practical applications.
\item We conduct a systematic evaluation of several widely used open-source datasets and data construction methods with the unified pipeline. Based on the results, we further distill practical data selection strategies and extract actionable insights into effective data construction, offering actionable guidance for cost-effective data curation and scalable model enhancement.
\item We propose several promising directions for future works, including RL-based data synthesis techniques, with the goal of encouraging further exploration into more scalable and effective training paradigms.
\end{itemize}





\section{Experimental Setup}
\label{sec:setup}

\begin{figure}[htb]
    \centering
    \includegraphics[width=0.9\linewidth]{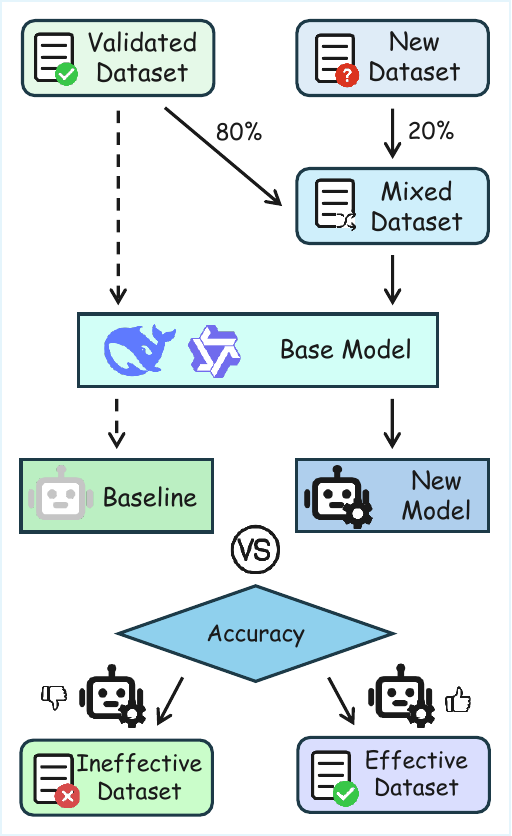}
    \caption{Evaluation pipeline for verifying the effectiveness of new datasets.}
    \label{fig-1}
\end{figure}

To ensure a consistent evaluation across datasets, we adopt a unified evaluation methodology. As shown in Figure \ref{fig-1}, we use \texttt{DeepSeek-V2-Lite} \citep{deepseekv2} or \texttt{Qwen2.5-3B} \citep{qwen2.5} as the base model and establish a baseline by training the base model on a large validated dataset consist of both code and mathematical data, covering both pretraining and SFT stages.

For evaluation, we follow the annealing method proposed in \citet{grattafiori2024llama3herdmodels} and \citet{cmrscalinglaw}, and train the base model on a mixture of the validated dataset and a new dataset under evaluation, where the new dataset is assigned a weight of 0.2 in the mixture. That is, 20\% of the training samples are drawn from the new dataset, while the remaining 80\% are sampled from the validated dataset. If the integration of data achieves better results than the baseline, we consider the evaluated dataset to be effective. 

This pipeline not only allows for a controlled assessment of the contribution brought by a specific dataset, but also aligns with evaluation practices commonly adopted in industrial scenarios. Such evaluation strategies are commonly adopted in production environments to ensure robustness and reproducibility. Therefore, our evaluation approach not only provides insights into the utility of different datasets, but also serves as a guideline for efficient data integration in real-world applications.

The benchmarks used for evaluation cover four types of tasks: common knowledge, logical reasoning, mathematical reasoning and coding ability. Accuracy is used as the metric for common knowledge, logical reasoning and mathematical reasoning tasks, while Pass@1 is used for coding tasks. Our primary focus is on improvements in mathematical reasoning, but we also pay attention to potential regressions in the model's performance on other capabilities. The benchmark used for evaluation are shown in Table \ref{tab:datasets}.

\begin{table}[htb]
\centering
\begin{adjustbox}{width=0.45\textwidth}
    \begin{tabular}{ c | l }
        \toprule  
        Domain & Dataset   \\
        \midrule
        \multirow{4}{*}{Knowledge} & MMLU \citep{mmlu}\\
        & MMLU-Pro \\
        & CMMLU \citep{cmmlu}\\
        & GPQA-Diamond \citep{gpqa}\\
        \midrule
        \multirow{3}{*}{Reasoning} & HellaSwag \citep{hellaswag}\\
        & BBH \citep{bbh} \\
        & DROP \citep{drop} \\
        \midrule
        \multirow{4}{*}{Math} & MATH \citep{math}\\
        & GSM8K \citep{gsm8k}\\
        & MathBench-a \citep{mathbench} \\
        & MathBench-t \\
        \midrule
        \multirow{2}{*}{Coding} & OpenAI-HumanEval \citep{humaneval}\\
        & Sanitized-MBPP \citep{mbpp} \\
        \bottomrule
    \end{tabular}
    \end{adjustbox}
    \caption{Benchmarks Used for Evaluation.}
    \label{tab:datasets}
\end{table}


%% file: selection.tex
\section{Data Selection}
\label{sec:selection}

\subsection{Evaluation of Open-Source Datasets}
\label{sec:open-source datasets}

Despite the wide availability of open-source datasets for mathematical reasoning, their quality remains highly inconsistent \citep{industrycorpus2, opencoder, limo, s1k}. In our practical experience, we find that simply adding more datasets does not necessarily lead to better model performance. In some cases, it can even jeopardize the overall performance. To better guide future data selection, we conduct a comparative analysis of both effective and ineffective datasets, and summarize key characteristics that distinguish "better data" from "more data".

Many open-source datasets are constructed by scraping web pages and filtering for relevant content. For example, IndustryCorpus2 \citep{industrycorpus2} aggregates and stratifies domain-specific data through rule-based and model-based filtering, and opc-fineweb-math-corpus \citep{opencoder} uses fastText to retrieve math-related pages from the Fineweb corpus. However, as our evaluation in Table \ref{tab:web-datasets} shows, the diverse nature of source websites makes simple cleaning strategies insufficient to remove noise thoroughly. This can degrade the overall dataset quality and hinder training effectiveness, highlighting that merely adding more data without careful curation does not necessarily improve model performance.

To address this limitation, recent works have explored an alternative paradigm: instead of collecting more data, they focus on curating small yet high-quality datasets through rigorous filtering and model-based selection. Notable examples include the LIMO \citep{limo} and s1K \citep{s1k} datasets, both of which are built through careful sampling from large candidate pools, followed by multi-stage filtering that emphasizes difficulty, diversity and reasoning depth. Furthermore, these datasets also leverage powerful reasoning models to generate detailed explanations, thereby enriching the data with high-quality reasoning trajectories. However, because of the limited data volume, the performance gains from these high-quality datasets remain modest. A promising direction is to scale up data construction by adopting similar methodologies. For example, OpenR1-Math-220K\footnote{\url{https://huggingface.co/datasets/open-r1/OpenR1-Math-220k}} proposes a dataset of over 220,000 samples and provides the model with more substantial performance improvements (e.g., +8.96 accuracy on the MATH dataset), according to the results in Tables \ref{tab:small-datasets} and \ref{tab:openr1}.

\subsection{Data Selection Strategies}
\label{sec:strategy}

Based on our evaluation of various open-source datasets, we summarize the following practical strategies for selecting effective datasets for mathematical reasoning tasks:

\begin{itemize}
[topsep=0pt,parsep=1pt,leftmargin=1.5em]
\item \textbf{Be cautious with data aggregated from diverse web sources, as more data are not always better.} Datasets collected through large-scale web scraping (e.g., IndustryCorpus2, opc-fineweb-math-corpus) often contain significant noise due to inconsistent formats. Without rigorous cleaning, such datasets may degrade the training performance.

\item \textbf{Prioritize datasets distilled by advanced reasoning models.} Data enriched with detailed reasoning steps or explanations generated by reasoning models (e.g., \texttt{DeepSeek-R1}, \texttt{QwQ-32B}) tend to provide better supervision signals and improve the model's own reasoning capabilities.

\item \textbf{Scale generation while maintaining data quality.} Instead of relying solely on manual curation, adopt scalable data generation methods that mimic the principles of high-quality datasets. These principles typically include leveraging LLMs to perform quality filtering and difficulty calibration. The success of OpenR1-Math-220K illustrates that principled large-scale generation is feasible and effective.
\end{itemize}

\begin{table*}[htbp]
  \centering
  \small
  \renewcommand\arraystretch{1.2}
  \begin{tabular}{c | p{8cm}}
  \toprule
  \textbf{Characteristic} & \textbf{Example} \\
  \midrule
  \multirow{2}{*}{\textbf{Provide clear definitions}} & To understand this concept more deeply, \textbf{let's first define} what it means ...\\
  \midrule
  \multirow{2}{*}{\textbf{Replace ambiguous expressions with terms}} &The study of \textbf{quadratic residues} and \textbf{nonresquares} modulo a prime number is ...\\
  \midrule
  \multirow{2}{*}{\textbf{Present formulas with LaTeX syntax}} & \multirow{2}{*}{We test \textbf{\textbackslash{}( x\^{}2 \textbackslash{}equiv 1 \textbackslash{}pmod\{83\} \textbackslash{})} and find that...}\\
  & \\
  \midrule
  \multirow{2}{*}{\textbf{Use examples instead of abstract explanations}} & Let's illustrate this process with a few examples. \textbf{Consider the number 2} ...
\\
\bottomrule
  \end{tabular}
  \caption{Examples of Pretraining Data Refinement}
  \label{tab:pretrain example}
\end{table*}

%% file: pretrain.tex
\section{Data Synthesis}
\subsection{Pretraining Data Refinement}
\label{sec:pretrain}

In the pretraining stage, the model builds fundamental language abilities and accumulates essential knowledge from large-scale data. While broad content coverage is necessary, simply adding more data does not guarantee better performance. It is equally important that pretraining data is structured in ways that enhance concept connections and facilitate deeper understanding.

Cosmopedia \citep{cosmopedia} serves as a representative example of pratraining data construction. They extract a wide range of valuable topics from educational sources such as Standford courses\footnote{\url{https://explorecourses.stanford.edu/search?q=all\%20courses}}, Khan Academy\footnote{\url{https://www.khanacademy.org/}}, OpenStax\footnote{\url{https://openstax.org/}}, WikiHow and other websites, and prompts an LLM to generate educational content tailored to different audiences and styles. 
By using \texttt{Mixtral-8x7B-Instruct-v0.1} \citep{mixtral}, they generate over 30 million files and 25 billion tokens of high-quality educational text covering diverse topics and styles.
Many other studies employ prompt engineering to construct pretraining data, all following a consistent approach: curating raw educational corpora from various sources through prompt design and using LLMs to refine them into more educational and well-structured content. 

We explore a similar approach in our experiments.
We begin by collecting books on various mathematical topics and apply OCR techniques\footnote{\url{https://mathpix.com/
}} to extract text from PDFs. After performing data denoising, we use resulting text as pretraining data, allowing model to learn knowledge through a simple next-token prediction task. Specifically, we focus on books related to intermediate algebra, resulting in 192 samples containing approximately 0.13 billion tokens. 
However, the result shows little improvement on standard math benchmarks, which suggests that merely scaling up data quantity alone is insufficient to ensure consistent performance gains.

Then, we make an attempt to replicate the methodology proposed by Cosmopedia \citep{cosmopedia}.
We use the math pretrain corpus previously collected and apply prompt engineering techniques to refine the corpus into a more educational format. Furthermore, we generate in-depth explanations based on the original content with \texttt{Qwen2.5-72B}. This process yields a dataset containing 760 million tokens, hereafter referred to as \textit{Math-Cosmo}. 
As a result, the performance of the model on the MATH and MathBench benchmarks are improved (e.g., +1.72 accuracy on the MATH dataset). The results of the experiment are presented in Table \ref{tab:math_cosmo}.

Moreover, we extend the Cosmopedia approach to Chinese mathematical pretraining data and generate high-quality samples that has been manually evaluated.
The prompt templates we used are shown in Appendix \ref{sec:prompt}.

\textbf{Insights:} When constructing pretraining data, simply collecting large amounts of data does not necessarily lead to performance gains. Instead, attention should be paid to how the data is presented to the model\citep{lu2023makespretrainedlanguagemodels}. Structuring it in a more interpretable format is an effective way to improve the model's reasoning capability. Examples are shown in Table \ref{tab:pretrain example}.

%% file: finetune.tex
\subsection{SFT Data Generation}
\label{sec:sft}

In the SFT stage, the model learns to follow instructions to complete specific tasks, thereby acquiring task-solving strategies and gradually connecting knowledge concepts gained during pretraining. This process often results in substantial improvement in reasoning and generalization abilities compared to the pretraining stage. However, simply increasing data quantity does not guarantee better performance. It is important to ensure the training data remains high-quality, diversity and difficulty.

A variety of studies have proposed different approaches to generate high-quality SFT datasets. S1 \citep{s1} employs two pretrained models to evaluate data, filtering out samples that can be correctly answered by the models, thereby ensuring the difficulty of the dataset. In addition, they classify the problems based on the Math Subject Classification system, randomly selecting samples from each domain while giving priority to data with longer reasoning steps, in order to enhance the diversity of the dataset. SynthLLM \citep{synthllm} leverages LLMs to extract topics and key concepts from documents, and combines them to generate new questions. Furthermore, it constructs a global concept graph based on the topic and concepts, and uses random walk algorithm to expand related knowledge concepts, to generate more complex and diverse questions. Dolphin-R1\footnote{\url{https://huggingface.co/datasets/cognitivecomputations/dolphin-r1}} directly employs larger models to generate data with reasoning steps. \citet{still-1} and \citet{omegaPRM} propose well-designed tree search algorithms to generate reasoning steps progressively.

In our experiment, we follow the methodology of Dophin-R1 and prompt \texttt{QwQ-32B} \citep{qwen2.5} to generate new answers for questions in NaturalReasoning \citep{naturalreasoning}. We then filter out instances with inconsistent answers, resulting in a new dataset, referred to as \textit{NaturalReasoning-QwQ}. As shown in table \ref{tab:natrualreasoning_qwq}, the experimental results clearly demonstrate that data distilled by reasoning model is of significantly higher quality than the original data, leading to substantial improvements in model performance (e.g., +1.92 accuracy on the MATH dataset). This reinforces the central insight that better data is more valuable than just more data. 

\begin{table}[htbp]
  \centering
  \scriptsize
  \renewcommand{\arraystretch}{1.2}  
  \setlength{\tabcolsep}{6pt}        
  \begin{adjustbox}{width=0.43\textwidth}
    \begin{tabular}{l | >{\centering\arraybackslash}p{1.5cm} >{\centering\arraybackslash}p{1.5cm}}
    \toprule
    \multirow{2}{*}{Benchmark} & \multicolumn{2}{c}{\textbf{NaturalReasoning-QwQ}} \\
     & \textbf{Score} & $\Delta$ \\
    \midrule
      MMLU               & 64.95 & –0.46 \\
      MMLU-Pro           & 14.58 &  1.39 \\
      CMMLU              & 78.06 &  0.15 \\
      GPQA-Diamond       & 32.32 & 0.5 \\
    \midrule
      HellaSwag          & 81.12 & –0.67 \\
      BBH                & 63.87 &  1.16 \\
      DROP               & \underline{63.48} &  0.64 \\
    \midrule
      MATH               & \underline{49.68} &  1.92 \\
      GSM8K              & 81.50 & –0.99 \\
      MathBench-a           & \underline{38.07} &  1.47 \\
      MathBench-t           & \underline{63.84} &  0.07 \\
    \midrule
      OpenAI-Humaneval   & 57.93 &  -0.61 \\
      Sanitized-MBPP     & 53.7 & –0.77 \\
    \bottomrule
  \end{tabular}
  \end{adjustbox}
  \caption{Performance of the model fine-tuned on the NaturalReasoning-QwQ dataset. "$\Delta$" denotes the difference in performance compared to the baseline model.}
  \label{tab:natrualreasoning_qwq}
\end{table}

To further examine this idea from a different perspective, we also explore a series of weakness-guided generation methods to acquire samples tailored to the model's limitations. More specifically, we first analyze the model's failure cases on the MATH dataset \citep{math} and define them as seed examples. Then, we use MathBERT \citep{mathbert} for embedding and FAISS \citep{faiss} to retrieve similar data from other datasets, in order to amplify their proportion in the training set. For each sample, we select the top 20 most semantically similar examples as candidates. 
Through this method, we collect over 75,000 examples comprising more than 82 million tokens, which we refer to as the \textit{math-retrieval dataset}. 
However, model gains little improvement from this dataset, especially on mathematical datasets.

We further prompt LLMs to generate data similar to these seed examples. Once new data is generated, we prompt the model to re-answer the same question. If the answers from both rounds are consistent, the data is considered valid. We directly apply this augmentation method to both seed examples and the \textit{math-retrieval} dataset, resulting in the \textit{math-weakness-augmented} dataset and \textit{math-retrieval-augmented} dataset. The detailed results are shown in Table \ref{tab:weakness-guided}, confirming that simply retrieving more data is less effective than applying high-quality augmentation strategies.

\begin{table*}[htbp]
  \centering
  \small
  \renewcommand\arraystretch{1.2}  
  \begin{tabular}{l | c c | c c | c c}
    \toprule
    \multirow{2}{*}{Benchmark} & \multicolumn{2}{c}{\textbf{Math-Retrieval}} & \multicolumn{2}{c}{\textbf{Math-Retrieval-Augmented}} & \multicolumn{2}{c}{\textbf{Math-Weakness-Retrieval}} \\
    & \textbf{Score} & $\Delta$ & \textbf{Score} & $\Delta$ & \textbf{Score} & $\Delta$\\
    \midrule
      MMLU               & 65.34 & –0.07 & 65.47 & 0.06 & 65.32 & -0.09\\
      MMLU-Pro           & 14.18 &  0.99 & 14.52 & 1.33 & 14.01 & 0.82 \\
      CMMLU              & 77.57 & -0.34 & 77.62 & -0.29 & 77.89 & -0.02 \\
      GPQA-Diamond       & 31.82 & 0 & 28.79 & -3.03 & 36.87 & 5.05\\
   \midrule
      HellaSwag          & 81.20 & –0.59 & 81.88 & 0.09 & 81.74 & -0.05 \\
      BBH                & 63.04 &  0.33 & 62.75 & 0.04 & 63.60 & 0.89 \\
      DROP        & 62.94 &  0.1 & 62.67 & -0.17 & \underline{63.45} & 0.61 \\
   \midrule
      MATH        & 46.06 &  -1.7 & \underline{51.48} & 3.72 & \underline{48.24} & 0.48 \\
      GSM8K       & 81.35 & –1.14 & \underline{82.64} & 0.15 & \underline{83.55} & 0.16 \\
      MathBench-a           & 37.4 & 0.8 & 36.4 & -0.2 & \underline{37.33} & 0.73\\
      MathBench-t           & 63.07 &  -0.7 & \underline{63.83} & 0.16 & \underline{63.88} & 0.11 \\
    \midrule
      OpenAI-Humaneval   & 63.41 &  4.87 & 56.71 & -1.83 & 58.54 & 0 \\
      Sanitized-MBPP     & 54.09 & –0.38 & 56.03 & 1.56 & 54.46 & -0.01\\
    \bottomrule
  \end{tabular}
  \caption{Performance of the model fine-tuned on datasets generated with weakness-guided methods. "$\Delta$" denotes the difference in performance compared to the baseline model.}
  \label{tab:weakness-guided}
\end{table*}

Furthermore, following \citet{scp}, we extract question-answer pairs from textbooks. In particular, we segment the content of the textbook into chunks based on chapters and sections, and use LLMs to extract question-answer pairs from each chunk, with each question paired with a corresponding solution. The generated data is then evaluated by LLMs to filter out low-quality samples. To ensure diversity, we further apply MinHash-based deduplication, resulting in the final dataset. We apply this approach to construct two datasets on intermediate algebra and calculus. Then we fine-tune the model separately on each dataset. Interestingly, two datasets constructed with the same methodology exhibit significant differences in the performance. This phenomenon may be related to the data mixing strategy, which warrants further investigation. Detailed performance of the models are shown in Appendix \ref{sec:textbook}.

\textbf{Insights:} When constructing SFT data, leveraging advanced models for data distillation or generating data tailored to the model's limitations can both effectively improve the model's performance, which highlights that better-curated data is more valuable than simply increasing data volume.

%% file: attempts.tex
\section{Unsuccessful Attempts}
\label{sec:attempts}

In our previous research, we also explored several approaches that did not yield satisfactory results. Here, we share these attempts in the hope of contributing to further exploration in this area.

There are some existing approaches constructing data through rule-based methods. \citet{fld} constructs multi-step reasoning data based on logical inference theorems (i.e., modus ponens), where the reasoning steps are logically valid but semantically meaningless.  \citet{xie2025memorizationlargelanguagemodels} generates Knights and Knaves puzzles at different difficulty levels with detailed reasoning steps. 
\citet{physicslanguagemodels} proposes a graph-based approach that allows models to learn dependencies between different objects. We follow this methodology and construct math word problems with different reasoning complexities based on the dependency graph and train the model on the resulting dataset. However, the improvements in reasoning ability brought by these methods are quite limited, especially in mathematics. The experiment shows that purely rule-based generation, without deeper quality control or task alignment, yields little benefit.

In addition, motivated by advanced Reasoning Large Language Models (RLLMs), we further combine the reasoning chains generated by the RLLMs (known as \textit{Long CoT} \citep{chen2025reasoningerasurveylong}) with the distilled data to construct the training dataset for our model. Still, the improvements in reasoning performance remain limited. We further categorize the data into different difficulty levels based on the length of reasoning steps, and evaluate the impact of difficulty on the model's reasoning ability. However, the results show no clear correlation between reasoning performance and data difficulty, again highlighting the difference between high-quality distillation and naive complexity scaling. 

\textbf{Insights:} Data constructed purely with rule-based methods generally provide limited benefit to the model, highlighting that lack of deeper structure may reduce its effectiveness. In addition, reasoning step length may not accurately reflect real difficulty, as no clear correlation is observed between longer reasoning chains and model performance. This further demonstrates that simply generating more data or increasing its apparent complexity does not guarantee better model reasoning.

%% file: future_work.tex
\section{Future Work}
\label{sec:future_work}

\begin{table*}[htbp]
  \centering
  \small
  \renewcommand{\arraystretch}{1.2}
  \begin{tabular}{l | c | c}
  \toprule
  \textbf{Dimension} & \textbf{"More Data"} & \textbf{"Better Data"}\\
  \midrule
  \textbf{Data Source} 
  & Noisy web-crawled 
  & Structured, interpretable \\
  \midrule
  \multirow{2}{*}{\textbf{Data Selection}}
  & \multirow{2}{*}{Broad aggregation without filtering}
  & Multi-stage filtering \\
  & & Model-based selection \\
  \midrule
  \multirow{2}{*}{\textbf{Data Synthesis}}
  & \multirow{2}{*}{Solely rule-based or logic-based} 
  & Distillation from advanced models\\
  & & Weakness-guided generation \\
  \midrule
  \textbf{Difficulty Control}
  & Partitioning by reasoning chain length
  & Fine-grained hierarchy with curriculum learning \\
  \midrule
  \textbf{Data Mixing}
  & Direct integration
  & Balanced, interference-aware \\
  \midrule
  \textbf{Outcome}
  & Inconsistent outcomes and conflicts across tasks 
  & Consistent improvements and cost-efficiency \\
\bottomrule
  \end{tabular}
  \caption{Summary of Key Findings: "More Data" vs. "Better Data" in Mathematical Reasoning.}
  \label{tab:conclusion}
\end{table*}

\subsection{RL-like Data Synthesis Methods}

Due to advanced development in reinforcement learning, we also propose several approaches for synthesizing pretraining and SFT data inspired by RL methods, such as \citet{zhang2025dadpodistributionawaredpodistilling} and \citet{du2025resureregularizingsupervisionunreliability}. Our motivation is to mimic the exploration process of RL in data generation to improve diversity, not merely to increase data quantity but to discover diverse, high-value examples that better improve the capabilities of models. Given the absence of experimental results, we present these ideas as exploratory directions. We believe such RL-like synthesis strategies hold  promising potential for generating better-curated data, and suggest several venues that merit further investigation.

\begin{itemize}
[topsep=0pt,parsep=0.5pt,leftmargin=1.5em]
    \item \textbf{multiple solution paths for the same problem} \space Generated data may include both correct and incorrect solutions, with varying levels of detail and diverse solution strategies.
    
    \item \textbf{reward-like signals} \space Labels such as quality-related metadata, annotations identifying reasoning flaws or suboptimal steps can be included to evaluate the generated data.
    
    \item \textbf{synthetic data with limited noise} \space Models should learn from both successes and failures. For instance, nearly correct solutions can be added to the training data along with explanations of their errors, enabling models to learn from mistakes.
    
    \item \textbf{hierarchical data training strategy} \space Gradually presenting the data to the model in order of increasing difficulty may lead to better performance improvements.
    
    \item \textbf{integration data from diverse models} \space The combination of synthesized data from models with different sizes or models possessing diverse domain knowledge may provide better diversity.
\end{itemize}

\subsection{Data Mixing Strategy}

In real-world scenarios, the model is required to handle multiple tasks rather than being confined to mathematical reasoning alone. As a result, it is common practice to mix data from different tasks during training. However, our findings show that indiscriminate data mixing can reduce the benefits that each individual datasets brings to the model, suggesting potential interference effects \citep{pike}. This suggests that simply adding more heterogeneous task data does not guarantee better performance. Therefore, to ensure optimal training outcomes, it is crucial to identify an appropriate data mixing strategy, especially for solutions designed for industrial environments.

\subsection{Curriculum Learning}

Some studies have proposed training strategies that progress from easy to hard, suggesting that such approaches can enhance the model's reasoning ability \citep{ji2025difficultyawarestagedreinforcementlearning}. However, our results show that simply partitioning data difficulty based on the number of reasoning steps is not effective, highlighting that more difficulty levels do not guarantee better outcomes if they are misaligned with actual complexity. Designing more fine-grained, task-relevant difficulty hierarchy and effectively incorporating the curriculum learning principles into the training process both warrant further investigation.

%% file: conclusion.tex
\section{Conclusion}
\label{sec:conclusion}

In this work, we systematically evaluate existing mathematical reasoning datasets and synthesis methods based on an industrial pipeline. Table \ref{tab:conclusion} summarizes our key findings in a direct comparison between "More Data" and "Better Data", which underscores that better-curated, high-quality data consistently outperforms simply increasing data volume. Overall, our study helps bridge the gap between theoretical research and real-world deployment, offering concrete guidance for cost-effective data curation and scalable model enhancement.

%% file: appendix.tex
\appendix
\section{Prompt for Chinese Mathematical Corpus}
\label{sec:prompt}

Figure \ref{fig:prompt} and \ref{fig:c_prompt} illustrates the prompt we use to construct Chinese mathematical pretraining data.

\begin{figure*}[!htbp]
\begin{tcolorbox}[colback=gray!10, colframe=gray!80]
\begin{lstlisting}
Below is a passage related to mathematics. You need to write an educational, detailed, and thorough article for both undergraduate and graduate students based on this passage.
"<EXTRACT>"
First, read the entire passage carefully, and list one by one the knowledge points mentioned but not explained in the passage-that is, the background knowledge the reader should have. Then provide explanations for each knowledge point. After that, proceed to write the main body of the article.
Before writing the main body, first consider the outline of the article. Then write the main body according to the outline.
In the main body, you need to elaborate extensively on the original passage, ensuring that all content from the original paragraphs is covered without omissions. Do not simply list concepts; instead, explore each concept in depth before moving on to the next.

Please pay special attention to the following points:
- Rigor: Ensure an in-depth analysis of each concept or section.
- Engagement: Use an academic, professional, and captivating writing style to stimulate the reader's interest.
- Format: Avoid using titles and introductory statements. Do not include images. After completing your reasoning, please output in the following format (do not add any other explanations):

==Background Knowledge==
(1) {Knowledge Point 1 and its explanation}
(2) {Knowledge Point 2 and its explanation}
...
==Main Body==
{Main body content}
==End==

\end{lstlisting}
\end{tcolorbox}
\caption{English version of the prompt for refining Chinese mathematical corpus.}
\label{fig:prompt}
\end{figure*}

\begin{figure*}[!htbp]
\begin{tcolorbox}[colback=gray!10, colframe=gray!80]
\begin{CJK*}{UTF8}{gbsn}
\begin{verbatim}
下面是一段数学相关篇章，你需要基于这段篇章为本科生以及研究生撰写一篇具有教育性的、详实且细致的文章。
"<EXTRACT>"
首先，通读全文，逐个列出篇章中涉及到但并未解释的知识点，即读者应具有的背景知识。并对知识点展开解释说明。然后进入正文撰写部分。
在撰写正文之前，先思考文章的大纲。然后按大纲撰写正文。
在正文部分，需要对原篇章进行详尽的展开，并确保覆盖原段落的所有内容，不能有遗漏。请勿简单罗列概念，而应在深入探讨每个概念后再转向下一个。

请重点关注以下几点：
严谨性：确保对每个概念或部分进行深入剖析。
吸引力：采用学术性、专业性且引人入胜的写作风格，以激发读者的兴趣。
格式：避免使用标题和介绍性语句。请勿使用图片。在结束思考后，请按如下格式输出（不要添加其他说明）：
==背景知识==
(1) {知识点1及其解释}
(2) {知识点2及其解释}
...
==正文==
{正文部分}
==结束==
\end{verbatim}
\end{CJK*}
\end{tcolorbox}
\caption{Chinese version of the prompt for refining Chinese mathematical corpus.}
\label{fig:c_prompt}
\end{figure*}

\section{Detailed Experimental Results}

\subsection{Open-source Datasets}
\label{sec:a_selection}

\begin{table}[htbp]
  \centering
  \small
  \renewcommand\arraystretch{1.2}  
  \begin{adjustbox}{width=0.45\textwidth}    
  \begin{tabular}{l | c c | c c }
    \toprule
    \multirow{2}{*}{Benchmark} & \multicolumn{2}{c}{\textbf{IC2-Math}} & \multicolumn{2}{c}{\textbf{opc-fineweb-math}} \\
    & \textbf{Score} & $\Delta$ & \textbf{Score} & $\Delta$ \\
    \midrule
      MMLU               & 63.28 & 1.29 & 60.95 & -1.04 \\
      MMLU-Pro           & - & - & - & - \\
      CMMLU              & 77.29 & 1.63 & 75.14 & -0.49\\
      GPQA-Diamond       & 33.84 & 8.59 & 26.26 & 1.01 \\
    \midrule
      HellaSwag          & - & - & - & -   \\
      BBH                & 56.14 & -6.28 & 61.45  & -0.97 \\
      DROP        & 64.03 & 0.74 & 62.85 & -0.44 \\
    \midrule
      MATH        & 36.74 & -3.7 & 32.9 & -0.14 \\
      GSM8K       & 72.86 & 2.81 & 70.89 & 0.84\\
      MathBench-a    & 32.27 & 1.94 & 29.33  & -1\\
      MathBench-t    & 51.12 & -3.46 & 52.41 & -2.17\\
    \midrule
      OpenAI-Humaneval   & 47.56 & -0.61 & 45.12  & -3.05\\
      Sanitized-MBPP     & 59.53 & -1.17 & 56.81 & -3.89\\
    \bottomrule
  \end{tabular}
  \end{adjustbox}
  \caption{Performance of the model trained on datasets collected through web scraping. "$\Delta$" denotes the difference in performance compared to the baseline model.}
  \label{tab:web-datasets}
\end{table}

\begin{table}[htbp]
  \centering
  \small
  \renewcommand\arraystretch{1.2}  
  \begin{adjustbox}{width=0.45\textwidth}    
  \begin{tabular}{l | c c | c c }
    \toprule
    \multirow{2}{*}{Benchmark} & \multicolumn{2}{c}{\textbf{LIMO}} & \multicolumn{2}{c}{\textbf{s1K}} \\
    & \textbf{Score} & $\Delta$ & \textbf{Score} & $\Delta$ \\
    \midrule
      MMLU               & 75.98 & -0.18 & 76.09 & -0.07 \\
      MMLU-Pro           & 52.41 & 0.23 & 52.77 & 0.59 \\
      CMMLU              & 84.65 & 0.08 & 84.54 & -0.03\\
      GPQA-Diamond       & 31.31 & 0 & 29.29 & -2.02 \\
    \midrule
      HellaSwag          & 91.35 & 0.06 & 91.31 & 0.02   \\
      BBH                & 74.07 & -0.25 & 73.87  & -0.45 \\
      DROP        & 72.18 & -0.5 & 73.06 & 0.38 \\
    \midrule
      MATH        & 54.9 & -0.28 & 54.52 & -0.66 \\
      GSM8K       & 86.58 & -0.15 & 89.01 & 2.28 \\
      MathBench-a    & 47.67 & -0.13 & 47.67  & -0.13\\
      MathBench-t    & 82.4 & 0.29 & 82.58 & 0.47\\
    \midrule
      OpenAI-Humaneval   & 50.61 & 0 & 48.78  & -1.83\\
      Sanitized-MBPP     & 65.37 & -1.17 & 64.98 & -1.56\\
    \bottomrule
  \end{tabular}
  \end{adjustbox}
  \caption{Performance of the model trained on samll yet high-quality datasets. "$\Delta$" denotes the difference in performance compared to the baseline model.}
  \label{tab:small-datasets}
\end{table}

\begin{table}[htbp]
  \centering
  \scriptsize
  \renewcommand{\arraystretch}{1.2}  
  \begin{adjustbox}{width=0.43\textwidth}
    \begin{tabular}{l | >{\centering\arraybackslash}p{1.5cm} >{\centering\arraybackslash}p{1.5cm}}
    \toprule
    \multirow{2}{*}{Benchmark} & \multicolumn{2}{c}{\textbf{OpenR1-Math}} \\
    & \textbf{Score}&  $\Delta$ \\
    \midrule
      MMLU               & 65.29 & –0.41 \\
      MMLU-Pro           & 12.94 & -0.57 \\
      CMMLU              & 78.32 & 0.29 \\
      GPQA-Diamond       & 32.83 & -2.02 \\
    \midrule  
      HellaSwag          & 80.43 & 0.37 \\
      BBH                & 63.19 & 0.26 \\
      DROP               & 62.36 & 0.42 \\
    \midrule
      MATH               & 52.2 & 8.96 \\
      GSM8K              & 81.65 & 0.91 \\
      MathBench-a           & 35.87 &  0.54 \\
      MathBench-t           & 60.54 & -2.22 \\
    \midrule
    OpenAI-Humaneval   & 50.61 &  -3.05 \\
    Sanitized-MBPP     & 48.25 &  -3.11 \\
    \bottomrule
  \end{tabular}
  \end{adjustbox}
  \caption{Performance of the model pretrained on the OpenR1-Math. "$\Delta$" denotes the difference in performance compared to the baseline model.}
  \label{tab:openr1}
\end{table}

\subsection{Pretraining Datasets}
\label{sec:a_pretrain}

Table \ref{tab:math_cosmo} illustrates the detailed performance of models pretrained on the \textit{Math-Cosmo} dataset. As shown in table, our synthesized dataset lead to improvements in the model's mathematical reasoning ability, thereby demonstrating the effectiveness of rewriting data into interpretable format.

\begin{table}[htbp]
  \centering
  \scriptsize
  \renewcommand{\arraystretch}{1.2}  
  \begin{adjustbox}{width=0.43\textwidth}
    \begin{tabular}{l | >{\centering\arraybackslash}p{1.5cm} >{\centering\arraybackslash}p{1.5cm}}
    \toprule
    \multirow{2}{*}{Benchmark}& \multicolumn{2}{c}{\textbf{Math-Cosmo}} \\
      & \textbf{Score} & $\Delta$ \\
    \midrule
      MMLU               & 65.02 & –0.39 \\
      MMLU-Pro           & 14.33 &  1.14 \\
      CMMLU              & 78.03 &  0.12 \\
      GPQA-Diamond       & 26.77 & –5.05 \\
    \midrule
      HellaSwag          & 80.53 & –1.26 \\
      BBH                & 64.07 &  1.36 \\
      DROP               & \underline{63.26} &  0.42 \\
    \midrule
      MATH               & \underline{49.48} &  1.72 \\
      GSM8K              & 82.41 & –0.08 \\
      MathBench-a           & \underline{37.93} &  1.33 \\
      MathBench-t           & \underline{63.80} &  0.03 \\
    \midrule
      OpenAI-Humaneval   & 59.76 &  1.22 \\
      Sanitized-MBPP     & 53.70 & –0.77 \\
    \bottomrule
  \end{tabular}
  \end{adjustbox}
  \caption{Performance of the model pretrained on the Math-Cosmo dataset. "$\Delta$" denotes the difference in performance compared to the baseline model.}
  \label{tab:math_cosmo}
\end{table}

\subsection{Textbook-based Datasets}
\label{sec:textbook}

Table \ref{tab:textbook} illustrates the performance of models fine-tuned on the textbook-based datasets. Two datasets constructed with the same method result in significantly different performance gains for the model. We believe this is related to the data mixing strategy.

\begin{table}[htbp]
  \centering
  \small
  \renewcommand\arraystretch{1.2}  
  \begin{adjustbox}{width=0.45\textwidth}
  \begin{tabular}{l | c c | c c}
    \toprule
    \multirow{2}{*}{\textbf{Benchmark}}& \multicolumn{2}{c}{\textbf{Intermediate Algebra-QA}} & \multicolumn{2}{c}{\textbf{Calculus-QA}}\\
    & \textbf{Score} & $\Delta$ & \textbf{Score} & $\Delta$ \\
    \midrule
      MMLU               & 65.30 &  -0.11 & 63.71 & -1.70 \\
      MMLU-Pro           & 15.04 &  1.85 & - & -\\
      CMMLU              & 77.82 & -0.09 & 78.39 & 0.48\\
      GPQA-Diamond       & 25.76 & -6.06 & 36.87 & 5.05 \\
    \midrule
      HellaSwag          & 81.67 & -0.12 & 79.26 & -2.53\\
      BBH                & 60.08 & -2.63 & 61.51 & -1.2\\
      DROP               & 62.91 &  0.07 & 62.53 & -0.31\\
    \midrule
      MATH               & 47.52 &  -0.24 & \underline{52.42} & 4.66\\
      GSM8K              & 82.18 &  -0.31 & \underline{84.53} & 2.04\\
      MathBench-a           & 36.67 &  0.07 & 33.67 & -2.93\\
      MathBench-t           & 63.18 &  -0.59 & 57.9 & -5.87\\
    \midrule
      OpenAI-Humaneval   & 62.8 &  4.26 & 59.15 & 0.61\\
      Sanitized-MBPP     & 55.64 & 1.17 & 55.64 & 1.17\\
    \bottomrule
  \end{tabular}
  \end{adjustbox}
  
  \caption{Performance of the model fine-tuned on the textbook-based dataset. "$\Delta$" denotes the difference in performance compared to the baseline model.}
  \label{tab:textbook}
\end{table}